\documentclass[sigconf,9pt]{acmart}
\usepackage[utf8]{inputenc}

\usepackage{enumitem}
\usepackage{booktabs}
\usepackage{comment}
\usepackage{graphicx}
\usepackage{multirow}

\usepackage{soul}

\definecolor{light_red}{RGB}{255, 204, 204}

\definecolor{crimson}{rgb}{0.86, 0.08, 0.24}

\newif\ifcomment

\commentfalse

\newif\ifacm
\acmfalse

\ifacm
\copyrightyear{2021} 
\acmYear{2021} 
\setcopyright{acmcopyright}\acmConference[EMDL'21 ]{5th International Workshop on Embedded and Mobile Deep Learning}{June 25, 2021}{Virtual, WI, USA}
\acmBooktitle{5th International Workshop on Embedded and Mobile Deep Learning (EMDL'21 ), June 25, 2021, Virtual, WI, USA}
\acmPrice{15.00}
\acmDOI{10.1145/3469116.3470012}
\acmISBN{978-1-4503-8597-8/21/06}
\else
\ccsdesc[500]{Computing methodologies~Neural networks}
\ccsdesc[300]{Human-centered computing~Ubiquitous and mobile computing systems and tools}
\keywords{Early-exit networks, Dynamic inference, Adaptive computing, Deep learning, Mobile systems}

\usepackage{background}
\backgroundsetup{angle=0,
    scale=1,
    color=black,
    firstpage=true,
    position=current page.north,
    hshift=0pt,
    vshift=-20pt,
    contents={\ifnum\value{page}=1 PREPRINT: Accepted at the 5th Annual International Workshop on Embedded and Mobile Deep Learning (EMDL), 2021 \else \fi}
}

\setcopyright{none}
\renewcommand\footnotetextcopyrightpermission[1]{} % removes footnote with conference information in first column

\fi

\ifcomment
\newcommand{\steve}[1]{\sethlcolor{cyan}\hl{[Stefanos: #1]}}
\newcommand{\alex}[1]{\sethlcolor{orange}\hl{[Alex: #1]}}
\newcommand{\cut}[1]{\sethlcolor{light_red}\hl{[#1]}}
\else
\newcommand{\steve}[1]{}
\newcommand{\alex}[1]{}
\newcommand{\cut}[1]{}
\fi
\title{Adaptive Inference through Early-Exit Networks:\\Design, Challenges and Directions}

\author[S. Laskaridis, A. Kouris, N. D. Lane]{\vspace{-6mm}{Stefanos Laskaridis$^\dagger$, Alexandros Kouris$^\dagger$, Nicholas D. Lane$^{\dagger\ddagger}$}}
	
\affiliation{\institution{$^\dagger$Samsung AI Center, Cambridge\hspace{+0.75cm}$^\ddagger$University of Cambridge}\country{}}

\email{
{stefanos.l, a.kouris, nic.lane}@samsung.com
}

\begin{document}

\vspace{-.1cm}
\begin{abstract}
DNNs are becoming less and less over-parametrised due to recent advances in efficient model design, through careful hand-crafted or NAS-based methods. Relying on the fact that not all inputs require the same amount of computation to yield a confident prediction, adaptive inference is gaining attention as a prominent approach for pushing the limits of efficient deployment. Particularly, early-exit networks comprise an emerging direction for tailoring the computation depth of each input sample at runtime, offering complementary performance gains to other efficiency optimisations. In this paper, we decompose the design methodology of early-exit networks to its key components and survey the recent advances in each one of them. We also position early-exiting against other efficient inference solutions and provide our insights on the current challenges and most promising future directions for research in the field. 
\end{abstract}

\maketitle
\ifacm
\else
\pagestyle{plain}
\fi
% \category{H.4}{Information Systems Applications}{Miscellaneous}
% %A category including the fourth, optional field follows...
% \category{D.2.8}{Software Engineering}{Metrics}[complexity measures, performance measures]

% \terms{Theory}

% \keywords{ACM proceedings, \LaTeX, text tagging}

\vspace{-.15cm}
\section{Introduction}
\vspace{-.1cm}
% \begin{itemize}[noitemsep]
%     \item Motivate the accuracy-latency trade-off i) as a way to fit DNNs on embedded devices, ii) as a way to dynamically allocate compute to more difficult examples, iii) as a way to dynamically scale compute based on system factors (battery, latency budget, etc.).
%     \item Fast overview of methods that can walk on the accuracy-latency curve and potentially comment on i) extra requirements (e.g. retraining) ii) distance to pareto.
%     \item From point (i) above, further motivate the importance of ``train-once, deploy everywhere'' paradigm for in-the-wild deployments. (Cite facebook paper for CPU-only).
% \end{itemize}

During the past years, there has been an unprecedented surge in the adoption of Deep Learning in various tasks, ranging from computer vision \cite{resnet} to Natural Language Processing (NLP) \cite{transformers} and from activity recognition \cite{dl_activity_recognition} to health monitoring \cite{emotionsense_www}. A common denominator and, undoubtedly, a key enabler for this trend has been the significant advances in hardware design \cite{embench_2019,ai_benchmark_2019} (e.g. GPU, ASIC/FPGA accelerators, SoCs) along with the abundance of available data, both enabling the training of deeper and larger models.
While the boundaries of accuracy are pushed year by year, DNNs often come with significant workload and memory requirements, which make their deployment on smaller devices cumbersome, be it smartphones or other mobile and embedded devices found in the wild. Equally important is the fact that the landscape of deployed devices is innately heterogeneous \cite{wu2019machine}, both in terms of capabilities (computational and memory) and budget (energy or thermal).

To this direction, there has been substantial research focusing on minimising the computational and memory requirements of such networks for efficient inference.
Such techniques include architectural, functional or representational optimisations in DNNs \cite{Wang2019}, aiming at faster forward propagation at a minimal cost. These include custom -- hand or NAS-tuned -- blocks \cite{mobilenets,efficientnet}, model weights sparsification and pruning \cite{DBLP:journals/corr/HanMD15} as well as low-precision representation and arithmetics \cite{rastegari2016xnor,ternary_nets}. 
Given there is no free lunch in Deep Learning, most of the aforementioned approaches trade off model accuracy for benefits in latency and memory consumption.
Moreover, while some of these approaches may work out-of-the-box, others do  require significant effort during or post training to maintain performance or to target different devices \cite{DBLP:journals/corr/HanMD15}.
% changes to the models, such as alternative operations (e.g. depth-wise separable convolutions, bottleneck blocks), sparsification and pruning of the model weights, low-precision representation and arithmetics as well as more efficient operation implementation (e.g. quantisation, Winograd convolutions, BNNs). 

\begin{figure}[t]
\centering
\vspace{-0.5em}
\includegraphics[width=0.8\columnwidth,trim={8cm 10cm 8cm 11.6cm},clip]{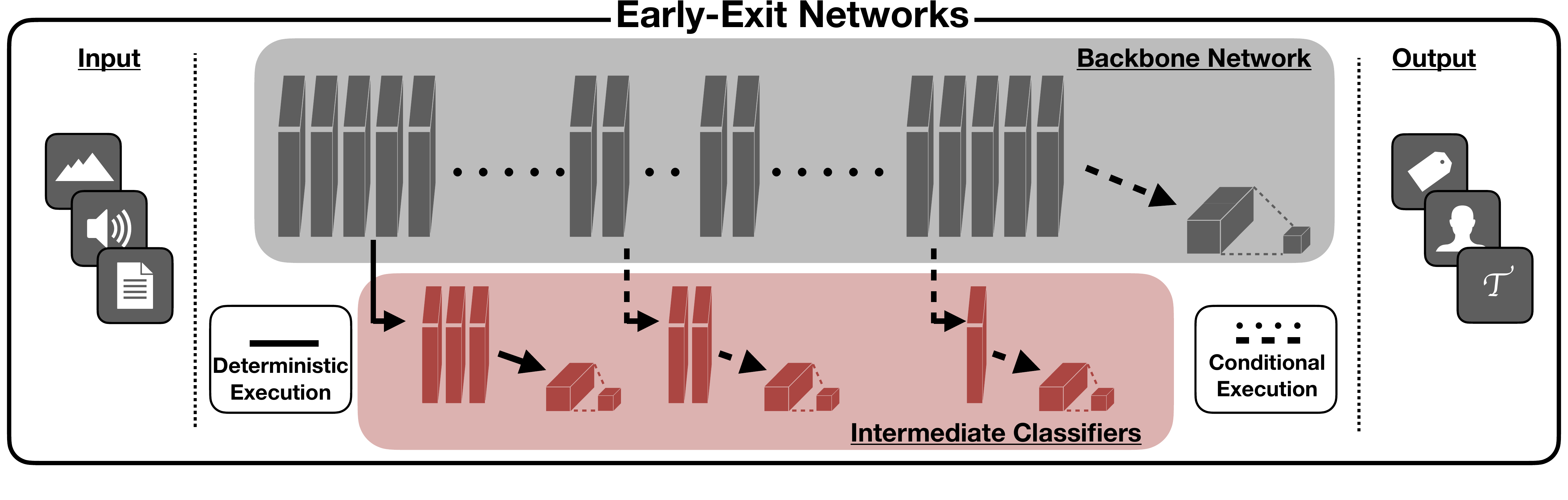}
\vspace{-0.6cm}
\caption{Early-exit network architecture}
\vspace{-0.6cm} %\vspace{-0.9cm}
\label{fig:ee-network}
\end{figure}

% \alex{Maybe we can motivate this inference scheme more as a complementary direction for pushing the limits of efficient inference (further than can be done by conventional methods)}
A complimentary family of solutions further pushing the efficiency envelope exploits the accuracy-latency trade-off at runtime, by adapting the inference graph \cite{nestdnn_2018,horvath2021fjord,wang2018skipnet,wu2018blockdrop} or selecting the appropriate model \cite{han2016mcdnn,Lee_2019} for the device, input sample or deadline at hand. This category includes early-exiting (EE) \cite{branchynet2016}.
Early-exit networks
% \footnote{Also known as progressive inference networks \cite{spinn2020mobicom}.} 
leverage the fact that not all input samples are equally difficult to process, and thus invest a variable amount of computation based on the input's difficulty and the DNN's prediction confidence; an approach resonating with the natural thinking mechanism of humans.
Specifically, early-exit networks consist of a backbone architecture, which has additional exit heads (or classifiers) along its depth (Fig.~\ref{fig:ee-network}). At inference time, when a sample propagates through the network, it flows through the backbone and each of the exits sequentially, and the result that satisfies a predetermined criterion (exit policy) is returned as the prediction output, circumventing the rest of the model. As a matter of fact, the exit policy can also reflect the target device capabilities and load and dynamically adapt the network to meet specific runtime requirements \cite{spinn2020mobicom,hapi2020iccad}.
% This process is also illustrated in Fig.~\ref{fig:early-exit-dnn}. 

\begin{figure*}[t]
\centering
\vspace{-1.6em}
\includegraphics[width=.6\textwidth,trim={0cm 8cm 0cm 8.5cm},clip]{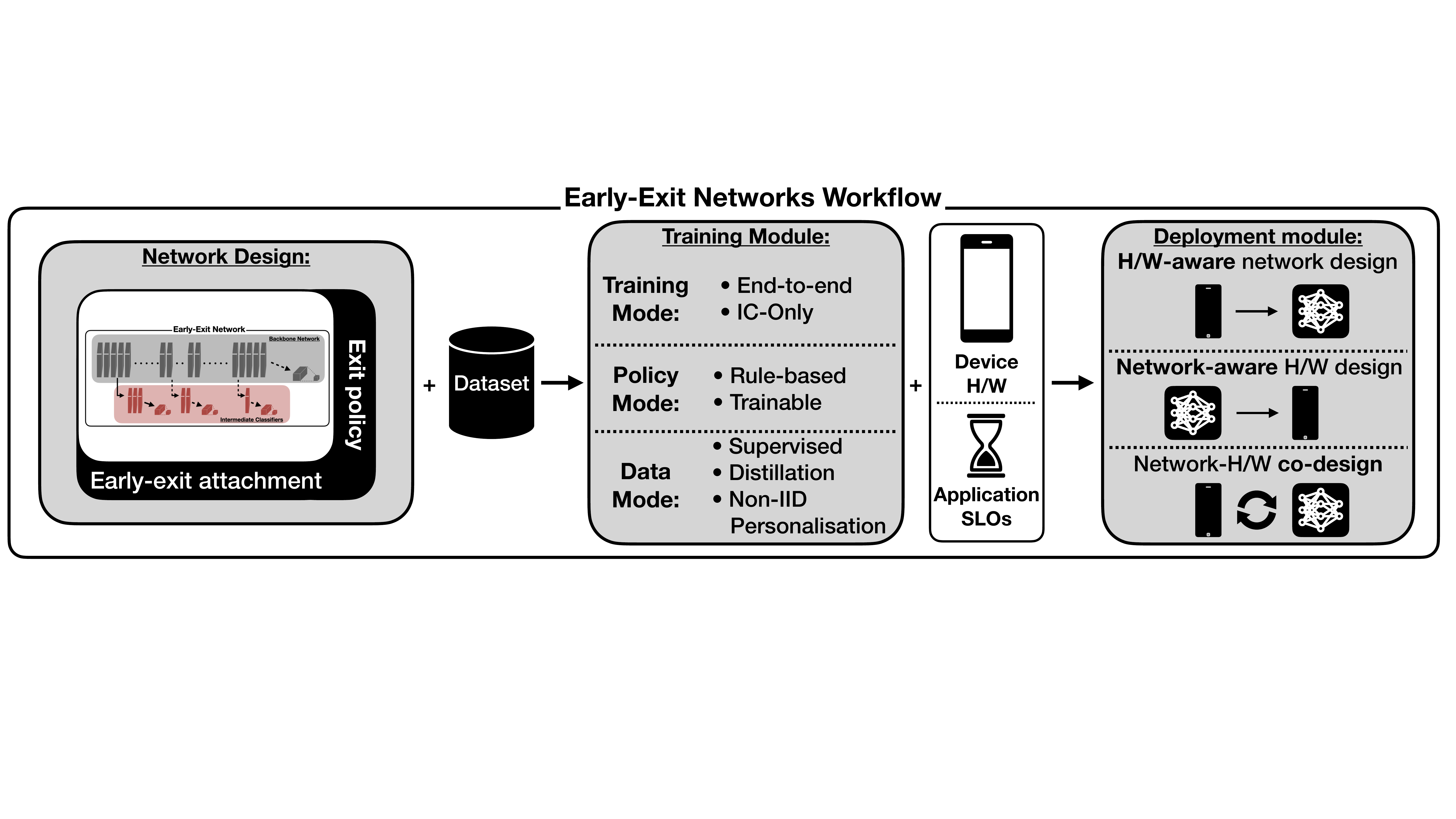}
\vspace{-1.1cm}
\caption{Early-exit networks workflow}
\vspace{-5mm}
\label{fig:ee-workflow}
\end{figure*}

Reaping the benefits of early exiting upon deployment, however, is not as trivial as jointly training a backbone network with randomly placed exits. One needs to carefully design the network and the training sequence of the exits relative to the backbone before choosing the exit policy for the deployment at hand. 
These decisions can be posed as a Design Space Exploration problem that can be efficiently traversed through a ``train-once, deploy-everywhere'' paradigm. This way, the training and deployment processes of early-exit networks can be detached from one another \cite{hapi2020iccad}.

This paper provides a thorough and up-to-date overview of the area of early-exit networks. Specifically, we first describe the typical architecture and major components of these networks across modalities. Next, we survey the state-of-the-art techniques
% for progressive inference 
and bring forward the traits that make such models a compelling solution. Last, we conclude by discussing current challenges in existing systems, the most promising avenues for future research and the impact of such approaches on the next generation of smart devices.

% Overall, the contributions of this work are:
% \begin{itemize}[noitemsep]
%     \item The cross-modal overview of early-exit inference techniques, their architecture, their training process and hyperparameter selection as well as the importance of the exit criterion for navigating the accuracy-latency tradeoff.
%     \item The discussion on current challenges in such networks, the most prominent avenues of future research and the impact on the next generation of mobile and embedded devices.
% \end{itemize}

\vspace{-.15cm}
\section{Early-Exit Networks}
% Introduce how early-exit networks work and what they aim to solve in a brief introduction. It would help to have an illustration here on how they work.

DNNs can be thought as complex feature extractors, which repre-
\newpage \noindent
sent inputs into an embedded space and classify samples based on the separability of classes in the hyperplane manifold. Typically, shallow layers extract low-level features, such as edges, whereas deeper ones build upon higher level semantics.
Under that framework, early exits can be thought as early decisions based on the shallower representations. The hypothesis behind their operation lays on the fact that such features on easier samples might be enough to offer the desired distinguishability between classes.

Several important decisions arise when designing, training and deploying such networks, however, as different designs affect the dynamics the network precision, performance and efficiency. In this and the following sections we go through the workflow of deploying an early-exit network (Fig.~\ref{fig:ee-workflow}).

\vspace{-2.9mm}
\subsection{Designing the architecture}
\vspace{-1mm}
% \begin{itemize}[noitemsep]
%     \item Number of exits
%     \item Position of exits
%     \item Architecture of exits (discuss per modality here?)
%     \item Manually design from scratch (e.g. MSDNet) or adapt to pre-existent backbone (e.g. everyone else)
% \end{itemize}

% \steve{Added some text here. Still needs refinement and discussion per modality I guess. I just don't want it to be too difficult to digest from the beginning. Also we need to carefully account for space I guess. I currently envision having approx. one figure section the least.}

\noindent
\textbf{Model \& Exit Architecture.}
Initially, one needs to pick the architecture of the early-exit model. There are largely two avenues followed in the literature: i) \textit{hand-tuned end-to-end designed networks} for early-exiting, such as MSDNet \cite{Huang2017}, and ii) \textit{vanilla backbone networks, enhanced with early exits} along their depth \cite{branchynet2016,hapi2020iccad,fang2020flexdnn,sdn_icml_2019}. 
This design choice is crucial as it later affects the capacity and the learning process of the network, with different architectures offering varying scalability potential and convergence dynamics. 

In the first case, networks are designed with progressive inference carved into their design. This means that the model and the architecture of its early exits are co-designed -- and potentially trained jointly. Such an approach allows for more degrees of freedom, but potentially restricts the design's performance across different circumstances and deployment scenarios, since this decision needs to be made early in the design process. For example, the existence of residual connections spanning across early exits can help generalisability of the network. On the other hand, some properties, such as maintaining multiple feature size representations, can prove detrimental in terms of model footprint \cite{Huang2017}.

On the other hand, when disentangling the backbone network's design from the early exits, one can have the flexibility of lazily selecting the architecture of the latter ones. Although this might not yield the best attainable accuracy, since the two components are not co-designed, it enables case-driven designs of early-exits that can be potentially trained separately to the main network and selected at deployment time \cite{hapi2020iccad}.

It is worth noting that early exits can adopt a uniform or non-uniform architecture, based on their placement. While the latter enlarges the design space of early-exit networks, it creates an interesting trade-off: The number (and type\footnote{Type can refer to the type of convolutions, such as regular vs. depthwise separable.}) of exit-specific layers accuracy vs. their overhead. While the adaptive and input-specific nature of early-exit networks is highly praised, when an early output does not meet the criteria for early-stopping, the runtime of the exit-specific layers are essentially an overhead to the inference computation. As such, the early exits need to be designed in comparison with the backbone network (i.e. relative cost) and with the exit policy at hand (i.e. frequency of paying that cost).

% TODO: Talk about NAS in future avenues for the architecture.

% creates an alternative two-staged training process, i.e. training the backbone separately to the early exits.

\noindent
\textbf{Number \& Position of Early-exits.}
In parallel with the architecture, one also needs to select the number and positioning of early exits along the depth of the network. This decision not only affects the granularity of early results, but also the overall overhead of early-exiting compared to the vanilla single-exit inference.
Too densely placed early exits can yield an extreme overhead without justifying the gains achieved by the extra classifiers, whereas too sparse placements can offer large refinement period until the next output is available. Moreover, having too many early classifiers can negatively impact convergence when training end-to-end.
With respect to positioning a given number of early exits, they can be placed equidistantly or at variable distances across the depth of the network. The decision depends on the use-case, the exit rate and the accuracy of each early exit. It is worth noting that this inter-exit distance is not actual ``depth'', but can be quantified by means of FLOPs or parameters in the network.

\vspace{-3mm}
\subsection{Training the network}
\vspace{-1mm}
% \begin{itemize}[noitemsep]
%     \item End-to-end
%         \begin{itemize}[noitemsep]
%             \item Cost function
%             \item Hyperparams (e.g. learning rate)
%         \end{itemize}
%     \item IC-only
%     \item Distillation techniques
%         \begin{itemize}[noitemsep]
%             \item Cost function weights (distillation alpha)
%             \item Temperature
%             \item Cross-exit distillation
%             \item Feature-map distillation
%     \item Personalisation through early-exits
%         \end{itemize}
%         % \alex{cross-exit distillation, feature-map distillation etc could be of interest, although are not extensively studied in terms of early-exit networks}
%     \item Personalised early-exits
% \end{itemize}

After materialising its architecture, the early-exit model needs to be trained on given dataset.
As hinted, there are largely two ways to train early-exit networks: i) \textit{end-to-end (E2E)} and ii) \textit{intermediate classifiers (IC) only}. Each approach presents different trade-offs in terms of achieved accuracy vs. flexibility for target-specific adjustments. Here, we discuss these trade-offs along with orthogonal training techniques that can boost the overall accuracy of the model. % at hand.

\vspace{-2mm}
\subsubsection{End-to-end vs. IC-only training}

% \vspace{-4mm}
\noindent
\textbf{\newline End-to-end training.}
The approach comprises jointly training the network and early exits. Normally, a joint loss function is shaped which sums intermediate and the last output losses ($L_{task}^{(i)}$) in a weighted manner (Eq.~\ref{eq:e2e_loss}) and then backpropagates the signals to the respective parts of the network. While the achieved accuracy of this approach can be higher both for the intermediate ($y_{i<N}$) and the last exit ($y_N$), this is not guaranteed due to cross-talk between exits \cite{Li_2019_ICCV}. Concretely, the interplay of multiple backpropagation signals and the relative weighting ($w_i$) of the loss components \cite{hu2019learning} needs to be carefully designed, to enable the extraction of reusable features across exits. As such, while offering a higher potential, E2E training requires manual tuning of the loss function as well as co-design of the network architecture and the populated exits \cite{sdn_icml_2019}.

\vspace{-4mm}
\begin{equation}\label{eq:e2e_loss}
\vspace{-1.6mm}
    L_{e2e}(y_0, \dots, y_N, y) = \sum_{i=0}^N w_i * L_{task}^{(i)}(y_i,y)
\end{equation}

% One can devise different variants of end-to-end training, such as alternating between training the backbone and the exits between rounds or 

\noindent
\textbf{IC-only training.}
Alternatively, the backbone of the network and the early exits can be trained separately in two distinct phases. Initially, the backbone of the network, which may or may not be early-exit aware, is trained - or comes pretrained. In the subsequent phase, the backbone network is frozen\footnote{Meaning that the weights of this submodel are not updated through backpropagation.}, early-exits are attached at different points of the network and are trained separately (Eq.~\ref{eq:ic_loss}). This means that each exit is only fine-tuning its own layers and does not affect the convergence of the rest of the network. Therefore, the last exit is left intact, there is neither cross talk between classifiers nor need to hand-tune the loss function.
As such, more exit variants can be placed at arbitrary positions in the network and be trained in parallel, offering scalability in training while leaving the selection of exit heads for deployment time \cite{hapi2020iccad}. Thus, a \textit{``train-once, deploy-everywhere''} paradigm is shaped for multi-device deployment.
On the downside, this training approach is more restrictive in terms of degrees of freedom on the overall model changes, and thus can yield lower accuracy than an optimised jointly trained variant.

\vspace{-4mm}
\begin{equation}\label{eq:ic_loss}
\vspace{-2mm}
    L_{ic\text{-}only}^{(i)}(y_i, y) = L_{task}^{(i)}(y_i, y)
\end{equation}

\subsubsection{Training with distillation}

An ensuing question that arises from the aforementioned training schemes is whether the early-exits differ in essence from the last one and whether there is knowledge to be distilled between them. To this direction, there has been a series of work \cite{zhang2019your, Phuong_2019_ICCV, always_personal_hotmobile, liu-etal-2020-fastbert,Li_2019_ICCV} that employ knowledge distillation \cite{Hinton2015} in a self-supervised way to boost the performance of early classifiers. In such a setting, the student $i$ is typically an early exit and the teacher $j$ can be a subsequent or the last exit ($j\geq i$). As such, the loss function for each exit is shaped as depicted in Eq.~\ref{eq:distillation_loss} and two important hyperparameters emerge, to be picked at design time; namely the distillation \textit{temperature} ($T$) and the \textit{alpha} ($\alpha$). The temperature effectively controls how ``peaky'' the teacher softmax (soft labels) should be while the alpha parameter balances the learning objective between ground truth ($y$) and soft labels ($y_j$).

\vspace{-3mm}
\begin{equation}\label{eq:distillation_loss}
\vspace{-1.5mm}
    L_{distill}^{(i)}(y_i, y_j, y) = L_{task}^{(i)}(y_i, y) + \alpha L_{KL}(y_i, y_j, T)
\end{equation}

\subsubsection{Training personalised early-exits}

Hitherto, early exits have been trained for the same task uniformly across exits. However, when deploying a model in the wild, user data are realistically non-IID\footnote{Non Identically and Independently Distributed.} and may vary wildly from device to device. With this in mind, there has been a line of work \cite{reda_mm,always_personal_hotmobile} that personalises early exits on user data, while retaining the performance of the last exit in the source global domain. 
In \cite{always_personal_hotmobile}, this is effectively accomplished through IC-only training, where the backbone network is trained on a global dataset and early-exits are then trained on user-specific datasets in a supervised or self-supervised manner. In the latter case, data labels are obtained from the prediction of the last exit. Orthogonally, knowledge distillation can still be employed for distilling knowledge from the source domain to the personalised exits, by treating the last exit as the teacher \cite{reda_mm,always_personal_hotmobile}. 
% \steve{REDA is doing Domain Adaptation, so merging both into one section might not be ideal. To rethink.}    

\vspace{-2mm}
\subsection{Deploying the network}

% \begin{itemize}[noitemsep]
%     \item Subnet-based inference
%     \item Anytime \alex{progressive refinements, relates to SVD-based low rank approximation works in LSTMs etc}
%     \item Budgeted (and multi-model budgeted)
%     % \item Multi-model budgeted (?)
% \end{itemize}

At this stage, an early-exit network has been trained and ready to be deployed for inference on a target device. There are largely three inference modes for early exits, each relevant to different use cases:

\begin{itemize}[noitemsep,label={},leftmargin=*,topsep=1pt]
    \item \textbf{Subnet-based inference.} A single-exit submodel is selected (up to a specified early exit) and deployed on the target device. The main benefit here is the single training cycle for models of varying footprint, which, in turn, can target different devices or SLOs\footnote{Service Level Objective: e.g.~max latency or min accuracy.}.
    \item \textbf{Anytime adaptive inference.} An adaptive inference model is deployed and each sample exits (early) based on its difficulty, the confidence of the network prediction and potential app-specific SLOs. This mode offers progressive refinement of the result through early-exiting and latency gains for easier samples.
    \item \textbf{Budgeted adaptive inference.} Similar to anytime inference, but with throughput-driven budget. Concretely given a total latency budget, the goal is to maximise the throughput of correct predictions. This means that the model elastically spends more time on some examples at the expense of early-exiting on others.
\end{itemize}
% \alex{imo, here anytime and budgeted are not framed very clearly. We can discuss this further.}
% \alex{A more formal way to present would be to suggest that subnet-based is a extreme of anytime (always exit) and progressive refinement the opposite extreme (never exit). Input-dependent stands in the middle, conditioned by more complex exit criteria.}\steve{Will refine.}
Next, we are focusing on the last two use-cases and more specifically how the exit policy is shaped.

\vspace{-2mm}
\subsubsection{Deploying for adaptive inference}

% \begin{itemize}[noitemsep]
%         \item Exit-policy
%         \begin{itemize}[noitemsep]
%             \item Rule-based:
%             \begin{itemize}[noitemsep]
%                 \item Confidence-based:
%                 \begin{itemize}[noitemsep]
%                     \item Softmax top-1
%                     \item Softmax entropy
%                     \item Calibration of confidence
%                 \end{itemize}
%                 \item n-agreement
%                 \item Class means
%             \end{itemize}
%             \item Learnable Exit Policies \steve{I guess the training needs to be revised to include these.}
%                     \begin{itemize}[noitemsep]
%                         \item EpNet \cite{epnet}
%                         \item Differential Branching in Deep Networks for Fast Inference \cite{scardapane2020differentiable}
%                         \item Learning to Stop While Learning to Predict \cite{chen2020learning}
%                         \item (TODO: Search) Probabilistic learning \& uncertainty
%                     \end{itemize}
%         \end{itemize}
%         \item Monotonicity of accuracy and ``overthinking''
% \end{itemize}
% \alex{i.e. should we pick the latest, or the most confident prediction to time.  Is there any work that uses intermediate predictions as an ensemble ? Also iirc, forward propagation of shallow-exit predictions to deeper-ones has been studied as well. }

Exit policy is defined as the criterion upon which it is decided whether an input sample propagating through the network exits at a specified location or continues. Picking the appropriate depth to exit is important both for performance and to avoid ``overthinking\footnote{Overthinking refers to the non-monotonic accuracy of ICs; i.e.~later classifiers can misclassify a sample that was previously correctly classified.}'' \cite{sdn_icml_2019}. Overall, there are i)~\textit{rule-based} and ii)~\textit{learnable} exit policies.

% \steve{This is currently way to long and reads like related work. Needs to be refined.}
\noindent
\textbf{Rule-based early-exiting.} 
Most works in progressive inference have been employing the softmax of an exit to quantify the confidence of the network for a given prediction \cite{berestizshevsky2019dynamically}. On the one hand, we have approaches where the criterion is a threshold on the entropy of the softmax predictions \cite{branchynet2016}. Low entropy indicates similar probabilities across classes and thus a non-confident output whereas higher entropy hints towards a single peak result. On the other hand, other approaches use the top-1 softmax value as a quantification of confidence. 
% Naturally, there can be alternatives in this spectrum, integrating only some of the softmax values, such as best vs. second best.
An overarching critique for using confidence-based criteria, however, has been the need to manually define an arbitrary threshold, along with the overconfidence of certain deep models. Solutions include calibrating the softmax output values \cite{Guo2017} or moving to different exit schemes. 
% \steve{Also remember to talk about the overhead of calculating the softmax (probably in the next section).}
% 
Alternative rule-based exit policies include keeping per class statistics at each layer \cite{class_means}, calculating classifiers' \textit{trust scores} based on sample distances to a calibration set \cite{trust_score_neurips} or exiting after $n$ exits agree on the result \cite{bert_loses_patience_neurips2020}. 

% Other rule-based exit policies include have been proposed in various works \cite{class_means, trust_score_neurips, bert_loses_patience_neurips2020}. Specifically, the former \cite{class_means} tracks output \textit{class means} at every layer of the model and a sample exits early if the output of the layer is close to a class mean.
% % 
% Alternatively, \cite{trust_score_neurips} employs a calibration set to gauge a sample's trustworthiness and shapes the exit policy based on a \textit{trust score}, defined as the ratio of distances of the input sample to the nearest different class and the distance to the predicted class. \steve{This work is not really on early exits, but rather on confidence estimation.}
% % 
% Last, \cite{bert_loses_patience_neurips2020} defines the notion of \textit{patience} as an early-exit mechanism and stops inference computation when $n$ different exits agree on the result. This agreement can be on the same class for classification or the same range of values for regression tasks. Despite the cross-task applicability, it is evident that the exit can be at least $n$ classifiers away from the optimal, leading to an non-negligible overhead.

\noindent
\textbf{Learnable exit policies.}
Expectedly, one may wonder why not to learn network weights and the exit policy jointly. To this direction there has been work approaching the exit policy in differentiable \cite{chen2020learning, scardapane2020differentiable} and non-differentiable \cite{epnet} ways. In essence, instead of explicitly measuring the exit's confidence, the decision on whether to exit can be based on the feature maps of the exits themselves. The exit decision at a given classifier can be independent of the others (adhering to the Markov property) or can be modelled to also account for the outputs of adjacent exits.

\vspace{-.3cm}
\section{Early-Exits \& Target Hardware}
% \begin{itemize}[noitemsep]
%     \item Train-once, deploy-everywhere paradigm
%     \item Tune to specific hardware (i.e. HAPI)
%     \item Hardware/Software co-design
% \end{itemize}

Early-exiting not only provides an elegant way to dynamically allocate compute to samples based on their difficulty, but also an elastic way to scale computation based on the hardware at hand.
Although we have presented so far design, training and deployment as three distinct stages, these can, in fact, be co-designed and co-optimised for targeting different devices and SLOs.

First, a considerable benefit of early-exit networks, as aforementioned, is their ``train-once, deploy-everywhere'' paradigm. Essentially, this means that an overprovisioned network -- e.g.~a network with densely placed early-exits -- can be trained and then different parts of it be deployed according to the device's computational budget, memory capacity or energy envelope and application's latency, throughput or accuracy objectives. In essence, tweaking i)~the classifier architecture, ii)~the number and positioning of early-exits and iii)~exit-policy to the hardware at hand can be posed as a Design Space Exploration (DSE) problem with the goal of (co-)optimising latency, throughput, energy or accuracy given a set of restrictions, posed in the form of execution SLOs \cite{hapi2020iccad}. Accurately modelling this optimisation objective subject to the imposed restrictions is important for yielding efficient valid designs for the use-case at hand and shaping the Pareto front of optimal solutions.
% This can be modelled as an optimisation problem integrating the aforementioned objectives, subject to the aforementioned restrictions.
% Such a multi-objective cost function can be:

% \vspace{-6mm}
% \begin{align}
%     & \max\limits_{s}~A(s) - \mathrm{w}_{\text{lat}} \cdot L_{\text{hw}}(d, s) - \mathrm{w}_{\text{energy}} \cdot E_{\text{hw}}(d, s)
%     \label{eq:coopt}
%     \\ & \text{ s.t. } L_{\text{hw}}(d, s) \le \epsilon_\text{lat},
%     m(s) \le \epsilon_{\text{mem}}, 
%     L_{\text{hw}}(d, s) \le \epsilon_\text{energy} \nonumber
%     \label{eq:coopt_w_sla}
% \end{align}

% \noindent
% where for candidate design $s$ and benchmark vectors $d$ on a given target device, $A$ represents a function of the achieved accuracy of the early-exit (model,policy) on a calibration set, $L$ a function of its average latency and $E$ a function of its average energy, weighed by $\mathrm{w}_{*}$. $m$ represents the memory footprint of the design and $epsilon_{*}$ the SLO on latency, memory or energy at inference time.

Traversing this search space efficiently is important, especially since it needs to be done once per target device. Therefore, end-to-end training is usually avoided in favor of the more flexible IC-only approach.
It should be noted, though, that the search is run prior to deployment, and its cost is amortised over multiple inferences.

Instead of searching for the optimal network configuration for fixed hardware, another set of approaches is to design the hardware specifically for early-exit networks \cite{9032146, farhadi2019novel,kim2020low} or co-design the network and hardware for efficient progressive inference \cite{8843626,9020551}.

% \begin{itemize}[noitemsep]
%     \item H/W Design
%     \begin{itemize}[noitemsep]
%         \item A 0.22–0.89 mW Low-Power and Highly-Secure Always-On Face Recognition Processor With Adversarial Attack Prevention \cite{9032146}
%         \item A novel design of adaptive and hierarchical convolutional neural networks using partial reconfiguration on FPGA (also talks about gating). \cite{farhadi2019novel}
%         \item Low Cost Early Exit Decision Unit Design for CNN Accelerator \cite{kim2020low}
%     \end{itemize}
%     \item Co-design
%     \begin{itemize}[noitemsep]
%         \item Hardware-Software Co-design Approach for Deep Learning Inference (not really doing anything ee-specific hw-wise). \cite{8843626}
%         \item DynExit: A Dynamic Early-Exit Strategy for Deep Residual Networks \cite{9020551}
%     \end{itemize}
% \end{itemize}

\vspace{-.4cm}
\section{Adaptive Inference Landscape}
%Related work will contain a fast overview of alternative architectural designs (e.g. of the NestDNN type or Ordered Dropout) \alex{I guess this includes SkipNet, blockDrop etc ?} and techniques (i.e. pruning, quantisation) \alex{quantisation, channel-pruning and early-exits can be presented as 3 orthogonal directions for approximation in DNNs. Their interplay in terms of input dependent inference could be an important study direction (i.e. which samples are more resilient to quantisation ? does this mean that these samples can also be resilient to early-exiting ? would a quantised DNN be less resilient to early-exiting than the full-precision one ?}. Also not to forget cascades and their distinct aim. \alex{This could include a discussion about latency vs throughput optimisation, and privacy (availability of training data).} \\ \alex{Online Model selection could also be relevant (?)}

%Maybe a table of family of techniques and their pros and cons would be a nice illustration for this section. 

%Our goal in this section is not to convince the reader that EE are the only relevant technique that beats all the rest, but that it's a technique that its relevant due to its nature and it's here to stay. Much like pitching it to Nic per se.

\noindent
\textbf{Offline accuracy-latency trade-off.}
DNNs have been getting deeper and wider in their pursuit of state-of the art accuracy. However, such models still have to be deployable on devices in the wild. As such, optimising DNNs for efficient deployment has been an extremely active area of research. 
Approaches in the literature exploit various approximation and compression methods \cite{Wang2019} to reduce the footprint of these models, including quantisation of network weights and activations \cite{rastegari2016xnor,ternary_nets,gholami2021survey}
or weight sparsification and pruning \cite{deng2020model,liu2020pruning,DBLP:journals/corr/HanMD15}.
A common denominator amongst these techniques is that they inherently trade off latency or model size with accuracy. This trade-off is exploited offline in a device-agnostic or hardware-aware \cite{yang2018netadapt} manner.
Alongside, recent models tend to become less redundant, and thus more efficient, through careful hand-engineering \cite{mobilenets}, or automated NAS-based design \cite{efficientnet,liu2018darts}
% tan2019mnasnet
of their architecture. 
These approaches remain orthogonal to adaptive inference, thus offering complementary performance gains.

%\steve{@Alex, can you merge references here with the ones in the introduction?} \alex{Sure! Do you mean to add here all relevant citations from introduction, or to cite the same papers on both ?}\steve{Mainly the 2nd.}
% , by aiming to push the approximation as much as each sample can tolerate without harming its predictive accuracy.

%Talk about orthogonality and open question of how early-exiting works e.g. on (partially) quantised networks. [NOT DONE THE LAST PART]

%-> NetAdapt

\noindent
\textbf{Dynamic Networks.} Techniques in this family take advantage of the fact that different samples may take varying computation paths during inference \cite{liu2018dynamic}, either based on their intricacy or the capacity of the target device.
% Various approaches in the literature exploit the fact that ``easy" samples require less computation to yield a confident prediction, by customising the computation path on a per-sample basis, at runtime. 
Such methods include dynamically selecting specialised branches \cite{mullapudi2018hydranets}, skipping  \cite{wang2018skipnet,wu2018blockdrop,veit2018convolutional} or ``fractionally executing" (i.e. with reduced bitwidth) \cite{shen2020fractional} layers during inference, and dynamically pruning channels \cite{horvath2021fjord,nestdnn_2018,gao2018dynamic,lin2017runtime} or selecting filters \cite{yu2018slimmable, yu2019universally, chen2019you}. These approaches typically exploit trainable gating/routing components in the network architecture. This, however, complicates the training procedure and restricts post-training flexibility for efficient deployment on different hardware.
% in the sense that the model needs retraining for deployment on different hardware.

% In essence the model has to be re-trained in order to traverse the underlying accuracy-latency trade-off for adopting to different devices, applications and deployment scenarios. 

%-> Convolutional Networks with Adaptive Inference Graphs. 
%-> SkipNet: Learning Dynamic Routing in Convolutional Networks.
%-> BlockDrop: Dynamic Inference Paths in Residual Networks.

%->Dynamic Channel Pruning: Feature Boosting and Suppression.
%-> Channel Gating Neural Networks
%-> NestDNN: Resource- Aware Multi-Tenant On-Device Deep Learning for Continuous Mobile Vision.
%-> Runtime Neural Pruning.
%-> Ordered Dropout

\noindent
\textbf{Inference Offloading.}
Orthogonally, there has been a series of work on adaptive inference offloading, where part of the computational graph of a DNN is offloaded to a faster remote endpoint for accelerating inference to meet a stringent SLO \cite{Kang2017}. Some \cite{edgent_2020, spinn2020mobicom} even combine early-exit networks with offloading.

\noindent
\textbf{Model Selection \& Cascades}
More closely related to early-exiting come approaches that train a family of models with different latency-accuracy specs, all deployed on the target device. 
This is achieved by trading off precision \cite{kouris2018cascade}, resolution \cite{yang2020resolution} or model capacity \cite{han2016mcdnn,wang2017idk} %to different degrees 
to gain speed, or by incorporating efficient specialised models \cite{wei2019self}. At inference, the most appropriate model for each input is selected through various identification mechanisms \cite{Lee_2019,taylor2018adaptive}, or by structuring the model as a cascade and progressively propagating to more complex models until a criterion is met \cite{kouris2020throughput}. Although seemingly similar to early-exiting, ``hard" samples may propagate through numerous cascade stages without re-use of prior computation.
% This inefficiency can be partially alleviated through large batching, making these techniques most relevant to throughput optimised approaches. 

%Here we talk about a ``cousin'' approach, that of network cascades and talk briefly about the limitation of no prior reuse.

%-> Focus: Querying Large Video Datasets with Low Latency and Low Cost
%-> MobiSR: Efficient On-Device Super-Resolution Through Heterogeneous Mobile Processors
%->Adaptive Deep Learning Model Selection on Embedded Systems.
%->MCDNN: An Approximation-Based Execution Framework for Deep Stream Processing Under Resource Constraints.
%->Resolution Adaptive Networks for Efficient Inference
%->CascadeCNN

%==============================================
%\noindent\textbf{Iterative Refinement.}
%TODO: Consider iterative Refinement works? 
%->RefineNet: Iterative Refinement for Accurate Object Localization
%->ApproximateLSTM
%-> Laplacian Pyramid Reconstruction and Refinement for Semantic Segmentation (?) 
%-> ... 
%==============================================

\noindent
\textbf{Early-exiting.}
% Here we talk about early exiting techniques. A table would be very relevant here. We can further organise:
% \begin{itemize}[noitemsep]
%     \item Specific architectures per modality/task (e.g. MSDNet, FastBert, DeeBert, etc.)
%     \item More general DSEs like SDN and HAPI
%     \item Other training approaches (i.e. adversarial robustness).
% \end{itemize}
% 
Early-exiting has been applied for different intents and purposes. 
Initially, single early-exits were devised as a mechanism to assist during training \cite{szegedy2015going}, as a means of enhancing the feedback signal during backpropagation and to avoid the problem of vanishing gradients. In fact, these exits were dropped during inference. Since then, however, early-exits have proven to be a useful technique of adaptive inference and have been applied successfully to different modalities and for different tasks. These previously discussed techniques are concisely presented in Table~\ref{tab:ee_related_work} and organised by their optimisation goal, input modality and trained task.

\setlength{\tabcolsep}{2pt}
\begin{table*}[t]
    \vspace{-5mm}
    \caption{Work in early exiting.}
    \vspace{-4.5mm}
    \centering
    \scriptsize
    \resizebox{0.9\linewidth}{!}{
    \begin{tabular}{l l l l}
    % \begin{tabular}{p{3.2cm} p{2cm} p{6.7cm}} %{|l|p{1.3cm}|p{1.4cm}|p{1.6cm}|p{1.6cm}|p{1.6cm}|}
        \toprule
        \begin{tabular}{@{}c@{}} \textbf{Category} \\  \end{tabular} & \begin{tabular}{@{}c@{}} \textbf{Title} \\  \end{tabular} & \begin{tabular}{@{}c@{}} \textbf{Modality/Task} \\  \end{tabular} 
        & \begin{tabular}{@{}c@{}} \textbf{Description} \\ \end{tabular}
        \\ \midrule
        % \multicolumn{3}{c}{\textbf{Early-exit network-specific techniques}} \\
        \multirow{14}{10em}{\textbf{Early-exit network-specific techniques}} & 
        MSDNet \cite{Huang2017,Li_2019_ICCV}
        & Vision/Classification
        & Hand-tuned multi-scale EE-network. \\
        % & Li et al. \cite{Li_2019_ICCV}
        % & Vision/Classification
        % & MSDNet E2E training improvements via gradient scaling and distillation. \\
        & Not all pixels are equal \cite{8100167}
        & Vision/Segmentation
        & Pixel-level EE based on difficulty for semantic segmentation.\\
        & Phuong et al. \cite{Phuong_2019_ICCV}
        & Vision/Classification
        & Distillation-based EE from later to earlier MSDNet exits. \\
        & RBQE \cite{xing2020early} 
        & Vision/Enhancement
        & UNet-like network with EE, for Quality Enhancement.\\
        & The Right Tool for the Job \cite{schwartz-etal-2020-right}
        & NLP
        & Jointly trained EE on BERT. \\
        & DeeBERT \cite{xin-etal-2020-deebert}
        & NLP/GLUE
        & Jointly trained EE on Ro(BERT)a models.\\
        & FastBERT \cite{liu-etal-2020-fastbert}
        & NLP
        & Distillation-based EE on BERT models.\\
        & Depth-Adaptive Transformer \cite{Elbayad2020Depth-Adaptive}
        & NLP/BLEU
        & Transformer-based EE for translation.\\
        & Bert Loses Patience \cite{bert_loses_patience_neurips2020}
        & NLP/GLUE
        & Patience-based EE on (AL)BERT models.\\
        & Cascade Transformer \cite{soldaini-moschitti-2020-cascade}
        & NLP/QA
        & Transformer-based EE rankers for Question Answering. \\
        & MonoBERT \cite{xin-etal-2020-early}
        & IR/Document Ranking
        & Asymmetric EE BERT for efficient Document Ranking. \\
        & Chen et al. \cite{chen2020don}
        & Speech
        & Speech separation with EE-transformers.\\
        \midrule
        % \multicolumn{3}{c}{\textbf{Early-exiting network-agnostic techniques}} \\
        \multirow{9}{10em}{\textbf{Early-exiting network-agnostic techniques}}
        & CDLN \cite{panda2016conditional}
        & Vision/Classification 
        & Primary early-exit work based on linear classifiers.\\
        & BranchyNet \cite{branchynet2016}
        & Vision/Classification
        & Entropy-based fixed classifier EE-technique. \\
        & SDN \cite{sdn_icml_2019}
        & Vision/Classification
        & E2E \& IC-only overthinking-based training EE. \\
        & HAPI \cite{hapi2020iccad}
        & Vision/Classification
        & Hardware-aware design of EE-networks via DSE. \\
        & Edgent \cite{edgent_2020}
        & Vision/Classification
        & Submodel selection for offloading through EE training. \\
        & SPINN \cite{spinn2020mobicom}
        & Vision/Classification
        & Partial inference offloading of EE-networks. \\
        & FlexDNN \cite{fang2020flexdnn}
        & Vision/Classification
        & Footprint overhead-aware design of EE-networks.\\
        & DDI \cite{wang2020dual} 
        & Vision/Classification
        & Combines layer/channel skipping with early exiting. \\
        & MESS \cite{mess_kouris}
        & Vision/Segmentation
        & Image-level EE based on difficulty for semantic segmentation. \\
        \midrule
        % \multicolumn{3}{c}{\textbf{Variable label distributions}} \\
        \multirow{3}{10em}{\textbf{Variable label distributions}}
        & Bilal et al. \cite{bilal2017convolutional}
        & Vision/Classification
        & Hierarchy-aware ee-CNNs through confusion matrices. \\
        & Bonato et al. \cite{bonato2021class}
        & Vision/Classification
        & Class prioritisation on EE. \\
        & PersEPhonEE \cite{always_personal_hotmobile}
        & Vision/Classification
        & Personalised EE-networks under non-IID data \\
        \midrule
        % \multicolumn{3}{c}{\textbf{Learnable exit policies}} \\
        \multirow{3}{10em}{\textbf{Learnable exit policies}}
        & Scardapane et al. \cite{scardapane2020differentiable}
        & Vision/Classification
        & Differentiable jointly learned exit policy on EE-networks. \\
        & EpNet \cite{epnet}
        & Vision/Classification
        & Non-differentiable exit policy for EE-networks. \\
        & Chen et al. \cite{chen2020learning}
        & Vision/\{Classification, Denoising\}
        & Jointly learned variational exit policy for EE-networks.\\
        \midrule
        % \multicolumn{3}{c}{\textbf{Adversarial robustness}} \\
        \multirow{2}{10em}{\textbf{Adversarial robustness}}
        & Triple-wins \cite{Hu2020Triple}
        & Vision/Classification
        & Accuracy, robustness, efficiency multi-loss. \\
        & DeepSloth \cite{hong2021a}
        & Vision/Classification
        & Adversarial slowdown-enducing attack. \\
        \midrule
        % \multicolumn{3}{c}{\textbf{EE-H/W (co-)design}} \\
        \multirow{5}{10em}{\textbf{EE-H/W (co-)design}}
        & Kim et al. \cite{9032146}
        & Vision/Identification
        & Low-power \& robust EE model+H/W for continuous face recognition. \\
        & Farhadi et al. \cite{farhadi2019novel}
        & Vision/Classification
        & FPGA partial reconfiguration for progressive inference.\\
        & Kim et al. \cite{kim2020low}
        & Vision/Classification
        & Single-layer EE and H/W synthesis for thresholding. \\
        & Paul et al. \cite{8843626}
        & Vision/Classification
        & Efficient EE inference on FPGA.\\
        & DynExit \cite{9020551}
        & Vision/Classification
        & Trainable weights in joint EE loss and FPGA dployment.\\
        \bottomrule
    \end{tabular}
    }
    \vspace{-5mm}
    \label{tab:ee_related_work}
\end{table*}

\noindent
\textbf{Other surveys.} There have been certain previous surveys touching on the topic of early-exiting, either only briefly discussing it from the standpoint of dynamic inference networks \cite{han2021dynamic} or combining it with offloading \cite{matsubara2021split}. To the best of our knowledge, this is the first study that primarily focuses on early-exit networks and their design trade-offs across tasks, modalities and target hardware.

\vspace{-.4cm}
\section{Discussion \& Future Directions}
Having presented how early-exiting operates and what has been accomplished by prior work, here we discuss the main challenges and most prominent directions for future research in the field.

% Here we will talk about challenges and the directions we identify as most prominent for future research, in short and medium term. %we see future research being in the short and medium term.

\vspace{-4mm}
\subsection{Open Challenges.}
\noindent\textbf{Modalities.} A lot of research efforts in early exits have focused on the task of image classification through CNNs, and only most recently NLP through Transformer networks. However, a large variety of models (e.g.~RNN, GAN, seq2seq, VAE) are deployed in the wild, addressing different tasks including object detection, semantic segmentation, regression, image captioning and many more. Such models come with their own set of challenges and require special handling on one or more of the core components of early-exit networks, which remain largely unexplored to date. 

\noindent\textbf{Early-exit Overhead.} Attaching early exits to a backbone network introduces a workload overhead for the samples where the exit at hand cannot yield a confident enough prediction. This overhead heavily depends on the architecture of the exit, its position in the network, the effectiveness of the exit policy and the task itself. Hence, instantiating the optimal configuration of early exits on a backbone network, which balances this overhead against the performance gains from exiting early, remains a challenging task. 

\noindent\textbf{Architectural Search Space} As previously established, there is a large interplay between the building blocks of early-exit networks. It is therefore desirable to co-optimise many design and configuration parameters such as exit number, placement, architecture and policy. This inflates the architectural search space and makes it computationally challenging to traverse, in search for optimal configurations. Structured or NAS-based approaches to explore this space could could provide an efficient solution to this end.

\noindent\textbf{Training Strategy.} Training early-exit networks is inherently challenging. Normally, early layers in DNNs extract lower-level appearance features, whereas deeper ones extract higher-level semantics, important for the respective task. In the realm of early-exiting, 
classifiers placed shallowly are natively pushing for the extraction of semantically strong features earlier in the network, which causes tension between the gradients of different exits, and may harm the overall accuracy in the case of e2e training. Conversely, IC-only trained early exits may lead to inferior accuracy or increased overheads. Developing a training strategy that can combine the best of both worlds remains an open question.

\noindent\textbf{Exit Policy.} Current hand-tuned exit policy treat prediction ``confidence" as a proxy to accuracy. Exit placement and training strategy may cause this predictions to become naturally under- or over-confident, leading to a probability distribution over layers that does not reflect the network's innate uncertainty \cite{Guo2017}. Developing exit strategies that better reflect the networks readiness-to-exit and potential ability-to-improve its prediction by propagating to the next exit is a challenging area of research. Additionally, it is important to allow such methodologies to remain adaptable post-training, in order to facilitate efficient deployment to use-cases with varying requirements and devices with different computational capabilities.

% \begin{itemize}[noitemsep]
%     \item Overhead-early exit balance -OK
%     \item Confidence as a proxy to accuracy; only measures a distribution over labels (against each other), not the network's innate probability estimation -OK
%     \begin{itemize}[noitemsep]
%         \item Lack of quantification of current solutions' distance to oracle.
%         \item Use smaller models (equal in FLOPs/params to early-exit submodels) as baselines.
%     \end{itemize}
%     \alex{Are the 2 items above future directions? not intergated to text currently.}
%     \item Early-exit on regression tasks -OK
%     \item Large search architectural space of heads, when exposing depth, type of layers, positioning and confidence to search. -OK
%     \item Tension between last and earlier exits in joint training. -OK
% \end{itemize}

%\alex{maybe emphasize also that most works have studied classification, and adaptation to other tasks (regression, detection, segmentation, language models etc) is not trivial.}

\vspace{-4mm}
\subsection{Additional future directions}

\noindent\textbf{Temporal-Awareness.} In video or mobile agent applications, strong correlations typically exist between temporally and spatially adjacent input samples, and hence their underlying predictive difficulty given previous predictions \cite{hu2020temporally}. There is therefore space to integrate historical or codec information to further optimise early-exiting.

\noindent\textbf{Hierarchical Inference.} In latency critical applications, having some higher-level actionable result from early on may be more important than waiting for an accurate finer-grained classification prediction. Early-exit networks can facilitate this paradigm, through hierarchical inference, with earlier exits providing more abstract -- and therefore easier to materialise -- predictions (e.g. ``dog"), before specialising their prediction in deeper exits (e.g. ``beagle'') \cite{bilal2017convolutional,zamir2017feedback}.

\noindent\textbf{Personalisation.} At deployment time, deep learning models often meet narrower distributions of input samples than what they have been originally trained for (i.e. detecting a single user or their relatively stationary environment).
% during video conferencing, or deploying a mobile robot in a consumers home).
In such cases, early exits can act as a self-acceleration mechanism, trained on the spot, e.g. through knowledge (self-)distillation from the final exit \cite{always_personal_hotmobile}, to maximise their performance by specialising to the target distribution. 

\noindent\textbf{Heterogeneous Federated Learning (FL).} In FL deployments, participating devices can have very heterogeneous computational and network capabilities \cite{horvath2021fjord}. As such, submodels of varying depth may be distributed to clients to train on their local dataset, thus improving participation and fairness while avoiding stragglers.

\noindent\textbf{Probabilistic Inference.} 
% A stream of works employs neural networks in combination with probabilistic models (e.g. Bayesian Neural Networks; BNNs).
% Such stochastic models, can explicitly capture the predictive uncertainty, across all stages of inference. 
Probabilistic models (e.g. Bayesian Neural Networks) have a native way of quantifying the predictive uncertainty of the network across all stages of inference \cite{gal2016dropout}.
This property of stochastic models can be exploited by the exit policy, rendering BNNs a natural fit for early exiting methodologies.

% % \noindent\textbf{Interpretability}

% \begin{itemize}[noitemsep]
%     \item NAS with early-exits \alex{Combine with Challenge? }
%     \item Personalisation with early-exits  -OK
%     \item Hierarchical early exit networks (i.e. superclasses of total labels being classified at shallower level)  -OK
%     \item Train with early-exits \alex{Combine with Challenge? }
%     \item Early-exits for heterogeneous FL -Stef?
%     \item Interpretability and early-exits (i.e. how do early-exits work in terms of feature extraction and why)
% \end{itemize}

% \section{Conclusions}
% \input{sections/n_conclusion}
%\end{document}  % This is where a 'short' article might terminate

%ACKNOWLEDGMENTS are optional
% \section{Acknowledgments}
% This section is optional; it is a location for you
% to acknowledge grants, funding, editing assistance and
% what have you.  In the present case, for example, the
% authors would like to thank Gerald Murray of ACM for
% his help in codifying this \textit{Author's Guide}
% and the \textbf{.cls} and \textbf{.tex} files that it describes.


\vspace{-2mm}
\bibliographystyle{ACM-Reference-Format}
\footnotesize
% \vspace{-3mm}
\bibliography{sigproc}  % sigproc.bib is the name of the Bibliography in this case
\end{document}